\documentclass[conference]{IEEEtran}

\usepackage{amssymb}
\usepackage{latexsym}
\usepackage{times}
\usepackage{soul}
\usepackage{url}
\usepackage{amsmath}
\usepackage{amsthm}
\usepackage{booktabs}
\usepackage{algorithm}
\usepackage{algorithmic}
\usepackage[table,xcdraw]{xcolor}
\usepackage{comment}
\usepackage{multirow}
\usepackage{amsfonts}
\usepackage{makecell}
\usepackage[colorlinks=true,linkcolor=blue]{hyperref}%

\usepackage{graphicx}
\usepackage[caption=false]{subfig}

\usepackage{url}
\usepackage{xcolor}
\definecolor{newcolor}{rgb}{.8,.349,.1}
\usepackage{times}
\usepackage{epsfig}
\usepackage{graphicx}
\usepackage{amsmath}
\usepackage{amssymb}
\usepackage{fixltx2e}
\usepackage[export]{adjustbox}

\usepackage{float}
\usepackage{pdfpages}

\usepackage{multirow}
\usepackage{booktabs}
\usepackage{multicol}

\definecolor{mygray}{gray}{.92}

\makeatletter
\def\thickhline{%
  \noalign{\ifnum0=`}\fi\hrule \@height \thickarrayrulewidth \futurelet
   \reserved@a\@xthickhline}
\def\@xthickhline{\ifx\reserved@a\thickhline
               \vskip\doublerulesep
               \vskip-\thickarrayrulewidth
             \fi
      \ifnum0=`{\fi}}
\makeatother

\newlength{\thickarrayrulewidth}
\title{Deep Learning for Video-based Person Re-Identification: A Survey}

\author{
Khawar Islam
\thanks{Khawar Islam is with FloppyDISK AI Research, Karachi, Pakistan}% <-this % stops a space
\thanks{Manuscript Submission XXXX XX, XXXX; revised August XX, XXXX.}}

\begin{document}

\maketitle

\begin{abstract}
Video-based person re-identification (video re-ID) has lately fascinated growing attention due to its broad practical applications in various areas, such as surveillance, smart city, and public safety. Nevertheless, video re-ID is quite difficult and is an ongoing stage due to numerous uncertain challenges such as viewpoint, occlusion, pose variation, and uncertain video sequence, etc. In the last couple of years, deep learning on video re-ID has continuously achieved surprising results on public datasets, with various approaches being developed to handle diverse problems in video re-ID. Compared to image-based re-ID, video re-ID is much more challenging and complex. To encourage future research and challenges, this first comprehensive paper introduces a review of up-to-date advancements in deep learning approaches for video re-ID. It broadly covers three important aspects, including brief video re-ID methods with their limitations, major milestones with technical challenges, and architectural design. It offers comparative performance analysis on various available datasets, guidance to improve video re-ID with valuable thoughts, and exciting research directions.
\end{abstract}


\section{Introduction}
With the endless efforts of computer vision and deep learning researchers, deep learning has accomplished exceptional success in person re-ID. In a few years, deep learning shows remarkable results in video re-ID and gives new birth to surveillance systems. With the rapid improvement in multimedia technology, video re-ID has gained much more attention in academia and the industrial sector over the last ten years  \cite{zheng2016person, nambiar2019gait}. The dominant reason for video re-ID popularity is to provide a wide range of services for public safety such as tracking each person with a unique ID, preventing crimes, behavior analysis, forensic investigation, etc. \cite{almasawa2019survey}. In intelligent video surveillance applications, video re-ID is defined as recognizing an individual person through various non-overlapping cameras from the huge number of gallery images \cite{chen2020learning}. It is one of the intriguing computer vision problems that are present among inter-camera variance challenges such as background clutter, occlusion, viewpoint, illumination changes, human pose variation and etc.
\par
\begin{table*}[t]
    \centering
     \setlength{\tabcolsep}{2.2mm}
    \caption{Comparison between existing survey papers and our survey paper. Our survey paper mainly focuses on video re-ID.}
    \scalebox{1.1}{
\begin{tabular}{lcccc} 
\rowcolor{mygray}
\thickhline
\hline
\textbf{Survey} & \textbf{Focus} & \textbf{Major Contribution} & \textbf{Video re-ID} & \textbf{Publication} \\ 
\hline
\cite{mazzon2012person} & Crowd & \begin{tabular}[c]{@{}c@{}}Drawbacks of existing approaches \\ Proposed simple knowledge-based method\end{tabular} & Partial & PRL \\ 
\hline
\cite{satta2013appearance} & \begin{tabular}[c]{@{}c@{}}Appearance\\Descriptors\end{tabular}  & \begin{tabular}[c]{@{}c@{}}Covers public datasets with current evaluation \\ Raised open and closed set re-ID scenarios\end{tabular} & Partial & arXiv \\ 
\hline
\cite{bedagkar2014survey} & Open-Closed & \begin{tabular}[c]{@{}c@{}}Highlighted public datasets with current evaluation \\ Raised open and closed set re-ID scenarios\end{tabular} & Partial & IVC \\ 
\hline

\cite{zheng2016person} & \begin{tabular}[c]{@{}c@{}}Image\\ \& Video\end{tabular}  & \begin{tabular}[c]{@{}c@{}}Discussed history and relationship of person re-ID \\ Hand-crafted and DL methods are reviewed \end{tabular} & Partial & arXiv \\ 
\hline

\cite{lavi2018survey} & re-ID  & \begin{tabular}[c]{@{}c@{}}Survey on deep neural networks techniques \\ Covers loss function and data augmentation \end{tabular} & X & arXiv \\ 
\hline

\cite{wang2018survey} & re-ID  & \begin{tabular}[c]{@{}c@{}} Traditional methods and architectural perspectives \\ CNN, RNN and GAN for
person Re-ID \end{tabular} & X &  CAAI-TIT \\ 
\hline

\cite{wu2019deep} & Image  & \begin{tabular}[c]{@{}c@{}} Survey of SOTA methods with feature designing \\Several results on ResNet and Inception \end{tabular} & P &  \begin{tabular}[c]{@{}c@{}} Neuro- \\ computing \end{tabular} \\ 
\hline

\cite{masson2019survey} & Image  & \begin{tabular}[c]{@{}c@{}} Extensively covered pruning methods, strategies \\ Performance evaluation on different datasets \end{tabular} & X &  JIVP \\ 
\hline

\cite{nambiar2019gait} & Gait  & \begin{tabular}[c]{@{}c@{}} Covered bio-metric details, pose analysis \\ Datasets and Multi-dimensional gait \end{tabular} & X &  ACM-CS \\ 
\hline

\cite{leng2019survey} & Open-world & \begin{tabular}[c]{@{}l@{}}Generalized open-world re-ID \\Specific application driven re-ID\end{tabular} & P & IEEE-TCSVT \\ 
\hline

\cite{wang2019beyond} & Heterogeneous & \begin{tabular}[c]{@{}c@{}}Focused on heterogeneous re-ID\\Problem of inter-modality discrepancies\end{tabular} & Partial & IJCAI \\ 
\hline

\cite{wang2020comprehensive} & \begin{tabular}[c]{@{}c@{}}Image\\ \& Video\end{tabular}  & \begin{tabular}[c]{@{}c@{}} Extensive review of previous Re-ID methods \\ Briefly discussed CNN, RNN and GAN \end{tabular} & X &  IEEE-Access \\ 
\hline

\cite{ye2021deep} & Image\&Video & \begin{tabular}[c]{@{}c@{}}Discussed closed-world and open-world re-ID\\Baseline for single-/cross-modality re-ID\end{tabular}  & Partial & IEEE-PAMI   \\
\hline

\cite{lin2021person} & \begin{tabular}[c]{@{}c@{}}Text \& \\Image\end{tabular} & \begin{tabular}[c]{@{}c@{}}Extensively reviewed person search methods \\ Feature learning and identity-driven methods \end{tabular}  & X & IJCAI  \\ 
\hline

\cite{lin2021unsupervised} & \begin{tabular}[c]{@{}c@{}}Image \& Video\end{tabular} & \begin{tabular}[c]{@{}c@{}}Extensively covered unsupervised methods\\Discussion about dataset and evaluation\\Peformance analysis and metrics\end{tabular}  & Partial & arXiV \\ 

\hline
\textbf{Ours} & \textbf{Video} & \begin{tabular}[c]{@{}c@{}} \textbf{Briefly discuss video re-ID methods} \\ \textbf{Discuss unique architectures, loss functions}\\ \textbf{Performance analysis of current methods} \end{tabular}  & \textbf{Full} & \textbf{CVIU} \\
\thickhline
\hline
\end{tabular}
}

    \label{tab:previous_papers}
\end{table*}
Video re-ID is an extended way of image-based person re-ID. Rather than comparing image pairs, pairs of video sequences are provided to the re-ID algorithm. The essential and important task of the video re-ID algorithm is to obtain temporal features from video sequences. Compare with image-based information, videos naturally comprise more information and evidence than individual images. Lately, numerous methods have been developed for video re-ID \cite{zhou2017see, zhang2017learning}. Most existing approaches emphasize extracting spatial and temporal features present in a video and then applying the re-ID algorithm to obtained features. In general, taking a video from different surveillance cameras like CCTV from different outside places. Then, detect persons in a video sequence and create a bounding box on it. Due to the high volume of data, it is difficult to draw manually bounding boxes and annotate each person's image for training. 
\par
Different studies \cite{ren2015faster,li2017scale,lan2018pedestrian} trained detectors to detect persons in a video sequence. Next, training a new re-ID model on highly noisy data based on previously annotated data. At last, giving query (probe) person image to re-ID model to find query person in a large set of candidate gallery \cite{ye2021deep}. The main role of video re-ID is to extract spatiotemporal features from video sequences. Some previous studies directly utilized person re-ID methods for images with some extension and applied for video. These approaches extract spatiotemporal information from each image independently by utilizing a recurrent neural network, feature aggregation function, and different pooling operations to obtain a frame-level information (e.g. appearance) representation. These above-mentioned techniques view different video frames with equal importance when needed frame-level features. However, these approaches extract abstract-level global features from the human body, while ignoring several local visual cues from a body such as a gait, hairs and etc. 
\par
Person re-ID in videos taken by multiple non-overlapping cameras which is a more practical implementation than images and achieves growing research trends \cite{wang2014person,zhou2017see}. In practical terms, videos captured from surveillance cameras with the involvement of pedestrians are the actual videos for person re-ID because these videos contain useful abundant information and spatial temporal features of a pedestrian that includes different human poses with diverse view angles. Nevertheless, recognizing discriminative portions of pedestrians against noisy data and extracting their features is an intriguing vision problem that is complicated for matching persons. Several video re-ID methods \cite{mclaughlin2017video,zhang2017learning} utilize CNN and RNN networks to extract spatio-temporal features from images and employ a pooling strategy to aggregate them. However, following these procedures, the task of matching persons becomes more sensitive when there are some noisy samples in data due to cluttered background or occlusion. While comparing two images of a person, each frame contributes equally to the matching task. For instance, if two persons are occupied with the same occluded object, the same appearance on occluded objects gives a false positive result in person re-ID.

\subsection{Contribution of this survey paper}
Most of the researchers focused on and surveyed traditional re-ID methods. Several survey papers covered conventional techniques including feature learning and distance learning, and some of them broadly covered deep learning techniques for re-ID. As far as our deep analysis, there is no survey paper discussing the recent video re-ID methods, novel loss functions, architectural designs, and approaches for video re-ID perspective. In this paper, we discuss comprehensive recent methods published in top-tier conferences and journals. In a nutshell, the contributions discussed in this survey paper are summarized as follows:
\begin{enumerate}
  \item To the best of our knowledge, this is the very first review paper to extensively cover deep learning methods for video re-ID instead of all types of person re-ID  compared with recent existing surveys \cite{ye2021deep,wu2019deep,almasawa2019survey}.
  \item We comprehensively cover deep learning techniques for video re-ID from multiple aspects, including global appearance methods, local part alignment methods, attention methods, graph methods, and transformer methods.
  \item This survey paper broadly covers architectural designs, novel loss functions, existing work, and the rapid progress of deep learning for video re-ID. Thus, it gives the readers to overlook the entire video re-ID work.
  \item Extensive comparison of top-ranking results on the benchmark datasets is performed. The development of video Re-ID and the challenges affecting video Re-ID systems are discussed, and a brief review and future discussion are given.
\end{enumerate}
\begin{figure*}[t]
    %\centering
     \includegraphics[width=\linewidth]{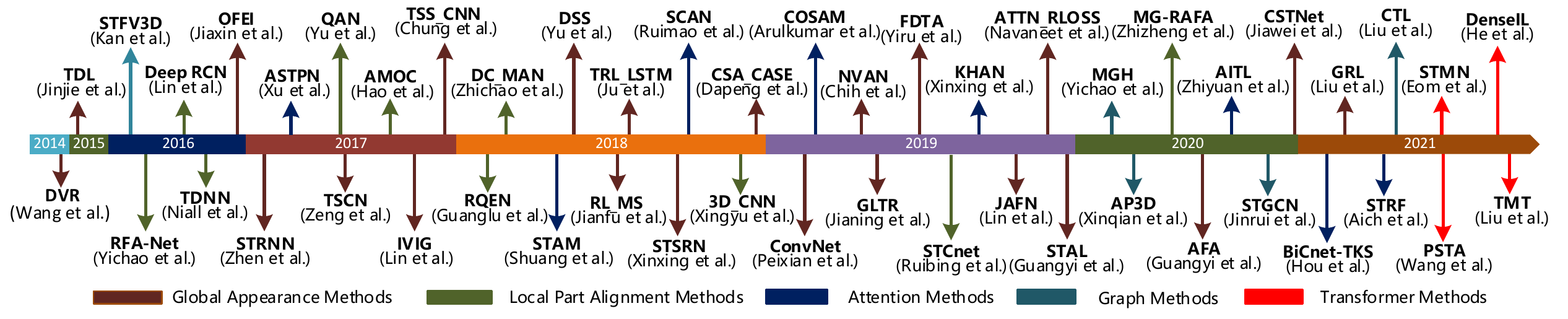}
    \caption{Timeline of the top-performing methods for video re-ID task.}
     \label{fig:Fig1}
\end{figure*}
\subsection{Review of existing survey papers}
Our recent survey offers a comprehensive and in-depth review of video re-ID and is distinct from previous and existing surveys and studies, as we broadly include the particular area of intelligent video surveillance and its practical usage. A detailed literature of previous survey articles aiming at person re-ID and its deep learning-based approaches is present, and some of them focused on open-world and close-world re-ID. Still, as far as we know, there is no previous paper that deeply focuses on video-based person re-ID from a practical point of view. We classify the previous and existing work of literature on person re-ID into five major categories named re-ID in the crowd, DL for re-ID, appearance descriptor for re-ID, opened the world, and closed re-ID. The comprehensive application scenario, relevant contributions, and special considered factors about past survey papers are described in Table \ref{tab:previous_papers}. The influential work on person re-ID applications is mentioned in \cite{zheng2016person,almasawa2019survey}. 
\par
\cite{mazzon2012person} presented a state-of-the-art (SOTA) re-ID framework for crowd applications and implementation of the practical framework in crowded scenes where people's movement captured from body appearance and comprehensively covered the discussion about person re-ID applications in terms of appearance features (color, texture, shape), association (distance, learning, optim) and calibration (color, spatial-temp). Similarly, Riccardo \cite{satta2013appearance}, provided a comprehensive overview of appearance descriptors and challenging issues i.e., illumination changes, partial occlusion, changes in color response and pose and viewpoint variations.
Furthermore, they covered global and local features with some “other cues”. Furthermore, the author in \cite{bedagkar2014survey} broadly discussed person re-ID problems, especially challenges related to the system level and component level. The authors discussed possible re-ID scenarios with comprehensively covered public datasets, evaluation metrics, and current work in re-ID. 
\par
One of the major and remarkable surveys \cite{zheng2016person} focused on each aspect of re-ID and connected with instance retrieval and image classification. A brief milestone and technical perspectives of image/video-based re-ID with hand-crafted methods were discussed. Traditional methods for person re-ID \cite{wang2018survey} were highlighted with further extended deep learning approaches such as CNN, RNN, and GAN to achieve person re-ID task and covered advantages and disadvantages. Similarly, the author in \cite{lavi2018survey} briefly discussed person re-ID from a video surveillance perspective and covered specific novel re-ID loss functions. The authors provided detailed re-ID approaches and divided them into i.e., recognized, verified deep model, distance learning metrics, feature learning, video-based person re-ID models, and data augmentation models. Further, they also conducted some experiments on the base model with several people re-ID methods \cite{wu2019deep}. In \cite{masson2019survey}, a detailed analysis of pruning techniques for compressing re-ID models is presented. To strengthen person the work, the authors further experimented with different pruning techniques and applied them to the deep siamese neural network. Their finding shows that pruning methods substantially decrease the number of parameters without decreasing accuracy. 
\par
Different from previous surveys, a gait-based \cite{nambiar2019gait} person re-ID has been discussed and highlighted various biometric regions in the human body i.e., hard biometrics including face identity, fingerprint, DNA, eye retina, and palmprint. Similarly, soft biometrics are related to body measurement, eye color, gait, and hair/beard/mustache. Particularly, the authors in \cite{almasawa2019survey} briefly discussed traditional and deep learning-based popular architectures and categories into image and video re-ID. Additionally, they compared to rank 1 results with SOTA methods and highlighted important challenges with future direction. Mostly person re-ID systems designed for closed-world settings, in \cite{leng2019survey}, the authors focused on open-world settings and discussed new trends of person re-ID in that area. They analyzed inconsistencies between open and closed-world applications and briefly discussed data-driven methods. Several specific
surveys \cite{mazzon2012person, nambiar2019gait, wang2019beyond} presented a depth literature review of some particular areas like heterogeneous re-ID \cite{wang2019beyond}, the author studied the concept of Hetero re-ID. They provided a comprehensive literature review in infrared images, low-resolution images, text, and sketches.
Afterward, the authors analyzed various datasets with evaluation metrics by giving future insights and providing new challenges areas in Hetero re-ID.
\par
Recently, the author \cite{ye2021deep} conducted an extensive literature review of deep learning-based re-ID. Instead of focusing on an overview, they briefly covered limitations and advantages. The new AGW baseline is designed with a novel evaluation metric (mINP) for single and cross-modality re-ID tasks. However, the above survey papers of all presented covered person re-ID surveys do not focus on recent methods of VID re-ID and their solutions for intelligent video surveillance and practical applications. Precisely, we cover recent novel loss functions designed for video re-ID, architectural design, brief technical aspects of significant papers, and broadly discuss performance analysis with the most frequent datasets used for video-based re-ID. Several popular methods are illustrated in Fig. \ref{fig:Fig1}

\section{Video re-ID Methods}
This section discusses the feature representation learning approaches for video re-ID. We divide it into five main categories: a) Global Appearance Methods (\autoref{sec:Global_Feature}) b) Local Part Alignments Methods (\autoref{sec:Local_Feature}) c) Attention Methods (\autoref{sec:Attention_Methods}) d) Graphs Methods (\autoref{sec:Graph_Methods}) and e) Transformers Methods (\autoref{sec:Transformer_Methods}).

\subsection{Global Appearance Methods}
\label{sec:Global_Feature}
This class of methods extracts a single feature vector from a person's image without any supplementary information. Since person re-ID is originally applied for person retrieval problems \cite{zhang2020ordered}, learning global feature is often ignored in previous studies when incorporating existing DL approaches into the video re-ID domain. As a pioneering work, \cite{mclaughlin2016recurrent} introduces the first \textbf{Recurrent Deep Neural Network (RDNN)} architecture based on pooling and re-currency mechanism to combine all time-step data into a single feature vector.
\par
To compare different temporal modeling methods, \cite{gao2018revisiting} comprehensively study 3D ConvNET, RNN, temporal pooling, and temporal attention by fixed baseline architecture trained with triplet and softmax cross-entropy losses. \cite{fu2019sta} address large-scale video re-ID problem by introducing their \textbf{Spatial Temporal Attention (STA)} method. Rather than extracting direct frame-level clues by using average pooling, a 2D ST map is used to measure clip-level feature representation without any additional clues. Generally, features extracted from a single frame contain a lot of noise, illumination, occlusion, and different postures. This results in the loss of discriminative information (e.g., appearance and motion). \textbf{Refining Recurrent Unit (RRU)} \cite{liu2019spatial} recovers the missing parts with the help of motion context and appearance from the previous frame. 
\par
Another popular solution is to explicitly handle alignment problem corruption using occluded regions. \cite{li2018diversity} employs a unique diversity regularization expression formulated on Hellinger distance to verify the SA models which do not find similar body parts. \cite{zhao2019attribute} propose an attribute-based technique for feature re-weighting frame and disentanglement. Single frame features are divided into different categories of sub-features, and each category defines a specific semantic attribute. A two-stream network \cite{song2019extended} that jointly handle detailed and holistic features utilize an attention approach to extract feature at the global level. Another network captures local features from the video and enhances the discriminative ST features by combining these two features.
\par
Different from \cite{zhang2020multi}, a \textbf{Global-guided Reciprocal Learning (GRL)} framework \cite{liu2021watching} extracts fine-grained information in an image sequence. Based on local and global features, \textbf{Global-guided Correlation Estimation (GCE)} module generates feature correlation maps, locating low and high correlation regions to identify similar persons. Further, to handle multiple memory units and enhance temporal features, \textbf{Temporal Reciprocal Learning (TRL)} is constructed to gather specific clues. \cite{li2021local} improve the global appearance by jointly investigating global and local region alignments by considering inter-frame relations.
\subsection{Local Part Alignments Methods}
\label{sec:Local_Feature}
These methods extract local part/region that effectively prevents misalignment with other frames in a tracklet. Considering the persistent body structure with the combination of inconsistent body parts in terms of appearance, they are new to each other. The goal is to distinguish personal images based on visual similarity.
\par
To preserve structural relationship details, the \textbf{Structural Relationship Learning (SRL)} \cite{bao2019preserving} is proposed to extract structural relations in a refined and efficient way. SRL helps convolutional features to make the relation useful between regions and GCN. GCN allows learning the feature representations of the hidden layers which encode node features and local structural information of graph. Another popular solution is \textbf{Spatial-Temporal Completion network (STCnet)} \cite{hou2019vrstc}, a method explicitly handles partial occlusion by recovering the occluded part appearance. \textbf{Region-based Quality Estimation Network (RQEN)} \cite{song2018region} designs an end-to-end training technique with gradient and learns the partial quality of each person image and aggregates complementary partial details of video frames in a sequence. 
\par
\begin{figure*}[t]
    \centering
    \includegraphics[width=\linewidth]{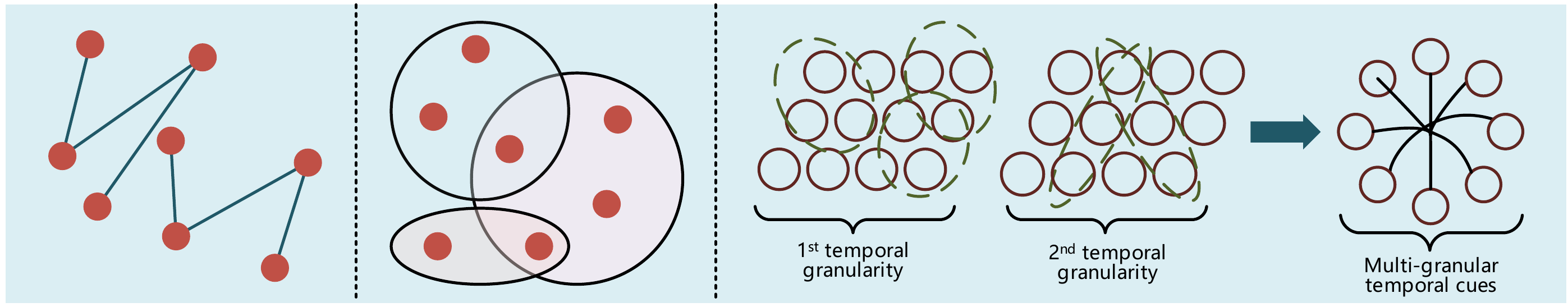}
    \caption{Illustration of simple graph each line connected with two vertices. In a hypergraph, each edge is connected with more than two vertices. In a multi-granularity graph, each node models specific spatial granularity, and each hypergraph is connected with multiple nodes.}
    \label{fig:hypergraph}
\end{figure*}
Different from previous methods, they utilize erasing techniques to penalize regularized terms during network training to prevent over-fitting. \cite{hou2020temporal} capture complementary affinities from video frames using an erasing strategy during training and testing. Based on the activated parts of previous frames, this approach erases the regions of each frame which ensures the frame concentrate on a new human part. To extract fine-grained cues, \textbf{Multi-Granularity Reference aided Attentive Feature Aggregation (MG-RAFA)} is proposed in \cite{zhang2020multi} to jointly handle Spatio-temporal features. Semantic hierarchy is considered for each node/position from a global point of view. For the position of each feature, local affinities are utilized with reference to feature nodes which provide the global structural and appearance information to support different weights to local features. \cite{li2021local} considers a holistic feature for visual similarity of video frames while focusing on the quality that allows the recovery of misaligned parts.

\subsection{Attention Methods}
\label{sec:Attention_Methods}
These methods usually ignore dissimilar pixels in training and prediction, employing similar pixels to make computational-friendly networks.
\par
\cite{song2018mask} introduce a mask-guided network, where binary masks are used to coexist with corresponding person images to decrease background clutter. Similar to the prior work, \cite{subramaniam2019co}, CO-Segmentation approaches have shown remarkable improvements in video re-ID over different baselines by integrating a \textbf{Cosegmentation-based Attention (COSAM) } \cite{subramaniam2021co} block among different layers in CNN networks. These CO-segmentation methods are able to extract unique features between person images and use them for channel and spatial-wise attention. In a different work in video re-ID, \cite{chen2019spatial} learn spatial-temporal features and calculate an attention score map to specify the quality of different components of a person.
\par
In real-world applications, the motion patterns of humans are the dominant part of re-ID. The Flow Guided-Attention network \cite{kiran2020flowguided} is designed to fuse images and sequence of optical flow using CNN feature extractor which allows encoding of temporal data among spatial appearance information. The Flow Guided-Attention depends on joint SA between optical flow and features to take out unique features among them. Additionally, an approach to aggregate features is proposed for longer input streams for improved representation of video sequences.
\par
Several studies focus on multi-grained and multi-attention approaches to concentrate on important parts of the human body. \cite{hu2020concentrated} introduce \textbf{Concentrated Multi-grained Multi-Attention Network (CMMANet)}, multi-attention blocks are proposed to obtain multi-grained details by processing intermediate multi-scale features. Moreover, multiple-attention sub-modules in multi-attention blocks can automatically discover multiple discriminative regions in the frame sequence. Relevant to multi-branch networks, \cite{hou2021bicnet} propose an innovative and computational-friendly video re-ID network that differs from the existing frameworks. \textbf{Bilateral Complementary Network (BiCnet)} preserves spatial features from the original image and down-sampling approach to broaden receptive fields and \textbf{Temporal Kernel Selection (TKS)} module captures the temporal relationship of videos. Different from previous studies, \cite{chen2020learning} introduces an end-to-end 3D framework to capture salient features of pedestrians in spatial-temporal domains. In this framework, salient 3D bins are selected with the help of two-stream networks and an RNN model to extract motion and appearance information.

\subsection{Graph Methods}
\label{sec:Graph_Methods}
After the remarkable success of the CNN model \cite{krizhevsky2012imagenet} in image understanding and reconstruction, academic and industrial researchers have focused on developing convolutional approaches for graph data. Recently, researchers combine re-ID methods with graph models and explore Video re-ID \cite{yan2016person}. \cite{cheng2018deep} develop a training network that jointly handles conventional triplet and contrastive losses through a joint laplacian form that can take complete benefit of ranking data and relationships between training samples. In \cite{shen2018person}, a novel unsupervised algorithm is formulated, which maps the ranking mechanism in the person re-ID method. Then, the formulation procedure is extended to  be able to utilize ranking results from multiple algorithms. Only matching scores produced by different algorithms can lead to consensus results. The key role of the person re-ID task is to efficiently calculate the visual resemblance among person images. Still, ongoing person re-ID methods usually calculate the similarity of different image pairs (investigated) and candidate lists separately whereas neglecting the association knowledge among various query-candidate pairs. 
\par
To solve the above problems, \textbf{Similarity-Guided Graph Neural Network (SGGNN)} \cite{chen2018group} propose to generate a graph to illustrate the pairwise associations between query and candidate pairs (nodes) and utilize these associations to provide up-to-date query candidate correlation features extracted from the image in an end-to-end manner. Most re-ID approaches emphasize local features for similarity matching. \cite{chen2018group} combine multiple person images to estimate the association between local relation and global relation in their \textbf{Conditional Random Field (CRF)}. The benefit of this model is to learn local resemblance metrics from image pairs whereas considering the dependencies of all images in a collection, shaping group similarities. \cite{yan2019learning} put more effort into person re-ID and employ context details. They first develop a contextual module called the instance expansion part, which emphasizes on relative attention part to find and purify beneficial context detail in the scene. One of the innovative works \cite{wu2020adaptive} for video re-ID is graph-based adaptive representation. Existing studies ignore part-based features, which contain temporal and spatial information. This approach allows the association between contextual information and relevant regional features such as feature affinity and poses alignment connection, to propose an adaptive structure-aware contagiousness graph. \cite{liu2021spatial} present \textbf{Correlation and Topology Learning (CTL)} method which generates robust and distinguish features. It captures features at multi-granularity levels and overcomes posing appearance problems.
\par 
Lately, hyper GNNs have attracted a lot of attention and achieved dominant results in various computer vision research fields such as person re-ID \cite{shen2018person}, action recognition \cite{wang2018videos} and image recognition \cite{chen2019multi}. These hypergraph algorithms develop pairwise relationships on the basis of object interest. In general, a hypergraph is a graph in which edges independently work and can join any considerable number of vertices. The illustration of hypergraph as shown in Fig. \ref{fig:hypergraph} (b) Conversely, as represented in Fig. \ref{fig:hypergraph} (a) where an edge exactly links with two vertices in a simple graph. In MG hypergraph, as represented in Fig. \ref{fig:hypergraph} (d) hypergraphs with distinct spatio granularities are built utilizing numerous stages of features like body part throughout the video frames. In every hypergraph stage, novel temporal granularities are taken by hyperedges which connect a type of nodes in a graph such as body part features around separate temporal scales. The first \textbf{Multi-Granular Hypergraph (MGH)} \cite{yan2020learning} hypergraph and innovative mutual information loss function are proposed to overcome the image retrieval problem. The MGH approach clearly supports multi-granular ST information from the frame sequence. Then, they propose an attention procedure to combine features presenting at the node level to obtain better discriminative graph representations. Remarkably, the proposed approach achieves 90\% rank-1 accuracy which is amongst the highest accuracy on the MARS dataset. Label estimation in graph matching is closely related to person re-ID problems in unsupervised learning. \cite{ye2019dynamic} present an unsupervised \textbf{Dynamic Graph Matching (DGM)} video re-ID approach to predict labels. This technique iterates the update process by utilizing a discriminative metric and correspondingly updated labels. 
\subsection{Transformer Methods}
\label{sec:Transformer_Methods}
Recently, transformer shows a great interest in the computer vision field, and self-attention-based methods are proposed to solve visual problems. Inspired by recent development, \cite{zhang2021spatiotemporal} put forward the first step, propose first \textbf{SpatioTemporal transformer (STT)}  and synthesize pre-training data strategy to reduce over-fitting for video re-ID task. In their network, the global module enables supplement to utilize the relation among patches from frames. To extract comprehensive features from videos, \cite{liu2021videoTTTrigeminal} further explore transformers and introduce \textbf{Trigeminal Transformers (TMT)} with robust novel feature extractor that jointly transform raw videos into S, T, and ST domains. To capture fine-grained features and aggregate in multi-view features, a self-view transformer is proposed to enhance single-view features and a cross-view transformer is used to combine multiple features. A \textbf{Duplex SpatioTemporal Filtering Network (DSFN)} \cite{zheng2021duplex} architecture is designed to extract static and dynamic data from frame sequences for video re-ID. To enhance the capability of kernels, sparse-orthogonal constraints are developed to broaden the difference in temporal features. To collaborate with a group of kernels, they add additional channels to assist and extract ST clues from distinct features. A Hybrid \textbf{Dense Interaction Learning (DenseIL)} framework is presented in \cite{he2021dense} which utilizes both CNN and Attention mechanism for video re-ID. DenseIL consists of a CNN-based encoder which is responsible to extract efficient discriminative spatial features and a DI-based decoder densely modeling the ST inherent interaction among frames.

\section{Novel Architectures}
% Different from existing architectures, \cite{jiang2021ssn3d} proposed a novel design to handle misalignment problem in video re-ID. \textbf{Self-Separated network (SSN)} provides an effective approach to deal with temporal and spatial variation of person body parts. SSN derived two-round classification approach, leads to the better training in pixel-wise and aggregated features.  An improved \textbf{Coarse-to-Fine Axial Attention Network (CF-AAN)} \cite{liu2021video} is designed with the help of Link and re-Detect block which can aligned noisy tracklet on the image level. This module not just decreased computational cost but also achieved promising results. Various video re-ID methods are still suffering in pose changes and person misalignment problems. To handle misalignment, \cite{zhang2021reference} proposed \textbf{Reference-Aided Part-Aligned (RAPA)} approach that focused on different parts of body and disentangle the discriminative features. \textbf{Reference Feature Learning (RFL)} pose-based module is provided to capture uniform standard for alignment. Aligning the body parts in intra-video, relations and attention-based \textbf{Part Feature Disentangling (PFD)} block is designed to locate and matching body parts through frames. 
Different from existing architectures, \cite{jiang2021ssn3d} propose a novel design to handle misalignment problems in video re-ID. \textbf{Self-Separated network (SSN)} provides an effective approach to deal with temporal and spatial variation of a person's body parts. SSN derives a two-round classification approach, leading to better training in pixel-wise and aggregated features.  The improved \textbf{Coarse-to-Fine Axial Attention Network (CF-AAN)} \cite{liu2021video} is designed with the help of Link and re-Detect block which can align noisy tracklist on the image level. This module not only decreases computational costs but also achieves promising results. Various video re-ID methods are still suffering from pose changes and personal misalignment problems. To handle misalignment, \cite{zhang2021reference} propose the \textbf{Reference-Aided Part-Aligned (RAPA)} that focuses on different parts of the body and disentangles the discriminative features. \textbf{Reference Feature Learning (RFL)} pose-based module is provided to capture uniform standards for alignment. Aligning the body parts in intra-video, relations, and attention-based \textbf{Part Feature Disentangling (PFD)} blocks are designed to locate and match body parts through frames. 
\par
% Most video re-ID methods focused on the important region of image, therefore, these methods can easily lose out fine-grained hints in image sequences. Different from previous studies, a novel GRL \cite{liu2021watching} framework is introduced along with reciprocal learning and correlation estimation. The GCE module creates the feature maps of local and global features that helps to locate the low region and high regions to identify similar person. Then, novel TRL approach is introduced to improve the high-correlation semantic information. \cite{gu2020appearance} proposed \textbf{Appearance Preserving 3D Convolution (AP3D)} and \textbf{Appearance-Preserving Module (APM)}, which aligned neighborhood feature maps in pixel-level, 3D ConvNets model temporal information on the basis of preserving the quality of visual appearance. It may be easier to aggregate AP3D with current 3DConNet by substituting prior 3D-Conv filters to AP3Ds. In video re-ID, person attributes and visual appearance are the most important and key to match each other, both features significantly contribute to the tracking of pedestrians. Novel TALNet \cite{liu2020temporal} is proposed to focus attribute-temporal learning by constructing branch network with the help of SA and temporal-semantic context.
\begin{table*}[t]
    \centering
    \caption{Training configuration of novel architectures. LR denotes learning rate and L represents loss}
    \scalebox{1.2}{
    \begin{tabular}{lcccccc} 
\rowcolor{mygray}
\thickhline
\hline
\multirow{1}{*}{\begin{tabular}[c]{@{}c@{}}\textbf{Ref }\end{tabular}} & \multirow{1}{*}

{\begin{tabular}[c]{@{}c@{}}\textbf{ Method}\end{tabular}} & \multirow{1}{*}

{\begin{tabular}[c]
{@{}c@{}}\textbf{Extractor}\end{tabular}} & \multirow{1}{*}{\begin{tabular}[c]{@{}c@{}}\textbf{L. Function}\end{tabular}} & \textbf{LR} & \textbf{Optimizer} & \textbf{Epochs}  \\ \hline
\cite{wang2014person}\textsubscript{ ECCV} & DVR & HOG3D & Hinge & — & — & — \\
\cite{karanam2015sparse}\textsubscript{ CVPR}& SRID & Schmid, Gabor filters & — & — & — & — 
\\
\cite{liu2015spatio}\textsubscript{ ICCV} & STFV3D & Fisher Vector & — & — & — & — \\
\cite{wu2016deep}\textsubscript{ ARXIV}  & Deep RCN & — & — & — & — & — 
\\
\cite{you2016top}\textsubscript{ CVPR} & TDL & \begin{tabular}[c]{@{}c@{}}HOG3D, Color \\ Histograms, LBP\end{tabular} & Hinge & — & — & — 
\\
\cite{chen2016person}\textsubscript{  IEEE-SRL} & OFEI & LBP & — & — & — & — 
\\
\cite{chen2016person}\textsubscript{  ECCV} & RFA-Net & LBP, HSV, Lab & Softmax & \begin{tabular}[c]{@{}c@{}}0.001 to \\ 0.0001\end{tabular} & — & 400 
\\
\cite{mclaughlin2016recurrent}\textsubscript{  CVPR} & CNN and RNN & Cross Entropy & 
 — &0.001 & SGD & 500 
\\
\cite{zhou2017see}\textsubscript{  CVPR} & JS-TRNN & TAM and SRM & Triplet & — & — & — 
\\
\cite{liu2017quality}\textsubscript{  CVPR} & QAN & — & Softmax and Triplet & — & — & — \\
\cite{xu2017jointly}\textsubscript{  ICCV} & ASTPN & — & CE and Hinge & 0.001 & SGD & 700 
\\
\cite{chung2017two}\textsubscript{  ICCV} & 2SCNN & CNN and RNN & Softmax & 0.001 & SGD & 1000 
\\
\cite{gao2021novel}\textsubscript{ ACM\_MM} & CMA & CNN+RNN & Softmax & 0.001 & SGD & 800 
\\
\thickhline
\end{tabular}
    }
    \label{tab:arch4}
\end{table*}
Most video re-ID methods focus on the important region of the image, therefore, these methods can easily lose out on fine-grained hints in image sequences. Different from previous studies, the novel GRL \cite{liu2021watching} framework is introduced along with reciprocal learning and correlation estimation. The GCE module creates the feature maps of local and global features that helps to locate the low regions and high regions to identify a similar person. Then, a novel TRL approach is introduced to improve the high-correlation semantic information. \cite{gu2020appearance} propose \textbf{Appearance Preserving 3D Convolution (AP3D)} and \textbf{Appearance-Preserving Module (APM)}, which align neighborhood feature maps in pixel-level. 3D ConvNets model temporal information on the basis of preserving the quality of visual appearance. It may be easier to aggregate AP3D with current 3DConNet by substituting prior 3D-Conv filters to AP3Ds. In video re-ID, personal attributes and visual appearance are key to matching identities, and both features significantly contribute to the tracking of pedestrians. Novel TALNet \cite{liu2020temporal} is proposed to focus on attribute-temporal learning by constructing a branch network with the help of SA and temporal-semantic context.

%Different from traditional triplet loss function, \cite{fang2021set} constructed a triplet loss which based on clip features. \cite{hu2020concentrated} introduced diversity loss. 
\section{Loss Functions}
Loss function plays a major and crucial role in discriminating the learned features. In general, the softmax loss separates the learned features rather than discriminates. The main goal of designing a person re-ID loss function is to enhance representation with an efficiency loss. We highlight several of the most influential loss functions for video re-ID.
\subsection{Attention and CL Loss}
\cite{pathak2020video} introduce CL centers online soft mining loss which utilizes center vectors from center loss as class label vector representations to crop out those frames that contain higher noise because it contains high variance compared to the original classifier weights. Additionally, they penalize the model by giving maximum attention scores to those frames that have randomly deleted patches. Those random erased frames are labeled as $1$ otherwise $0$ and N is the number of total frames.
%
%\vspace{-3pt}
\begin{equation}
    \begin{aligned} \mathrm{AL}=& \frac{1}{N} \sum_{i=1}^{N} \operatorname{label}(i) * \text { Attention }_{score}(i) \end{aligned}
\end{equation}
\subsection{Weighted Triple-Sequence Loss (WTSL)}
\cite{jiang2020weighted} explicitly encode frame-based image level information into video level features that can decrease the effect of outlier frame. Intra-class distance in WTSL makes similar videos closer and inter-class distance pushes dissimilar videos further apart. 
%
%\vspace{-3pt}
\begin{equation}
    \begin{aligned} \mathrm{L}_{W T S L}=& \sum_{i=1}^{N}\left[\left\|\mathrm{~F}_{a}^{i}-\mathrm{F}_{p}^{i}\right\|_{2}^{2}\right. \left.-\left\|\mathrm{F}_{a}^{i}-\mathrm{F}_{n}^{i}\right\|_{2}^{2}+\alpha\right]_{2} \end{aligned}
\end{equation}
where $\alpha$ represents margin, N is the number of triple-sequences and P represents person ID. The F\textsubscript{a} is a closer feature to its own class centroid and far away from other class centroids.
\subsection{Symbolic Triplet Loss (STL)}
\cite{aruna2020symbolic} propose STL which utilizes the Wasserstein metric to overcome the representation problem which allows obtaining the distance between feature vectors that are symbolic. 
%\vspace{-3pt}
\begin{equation}
    D_{w}\left(\psi_{i}, \psi_{j}\right)=\sum_{m=1}^{M} \sum_{t=1}^{T} \psi_{i m}{ }^{-1}(t)-\psi_{j m}{ }^{-1}(t)
\end{equation}
where $\psi_i$ and $\psi_j$ denote the distributions of multi-dimensional feature vectors at the i$^{th}$ and j$^{th}$. ${\psi_i}^{-1}(t)$ is the quantile function and M is the feature of each video.
\subsection{Weighted Contrastive Loss (WCL)}
\cite{wang2019deep} construct WCL by the combination of traditional contrastive loss. The purpose of this loss function is to allocate an appropriate weight for every proper image pair.
%
%
%\vspace{-3pt}
\begin{equation}
L_{W C L}(N)=\frac{1}{2} \frac{\sum_{\left(x_{i}, x_{j}\right) \in N} w_{i j}^{-} \max \left(0, \alpha-d_{i j}\right)^{2}}{\sum_{\left(x_{i}, x_{j}\right) \in N} w_{i j}^{-}}
\end{equation}
%\vspace{-3pt}
\begin{equation}
	\begin{aligned}
		&L_{WCL}(P,N) = (1 - \lambda) L_{WCL}(P) + \lambda L_{WCL}(N)
	\end{aligned}
\end{equation}
where hyperparameter $\lambda$ handles the contribution of both positive and negative sets towards final value of contrastive loss. 

% Please ignore this
\begin{comment}
\cite{xi2019fine} introduce feature mimicking loss to capture mid-level information because it contains more spatial information about re-ID. Semantically, high level features are more worthy but less descriptive of spatial data. Thus, mimicking loss jointly handles features and learn representations between low and high-level features.
%
\vspace{-3pt}
\begin{equation}
    \ell_{f m}=1-\frac{\mathbf{r}_{C o n v 4-6} \cdot \mathbf{r}_{C o n v 5-3}}{\left\|\mathbf{r}_{C o n v 4-6}\right\| \times\left\|\mathbf{r}_{C o n v 5-3}\right\|}
\end{equation}
\end{comment}
\begin{table*}[t]
    \centering
     \setlength{\tabcolsep}{5.5mm}
    \caption{Performance analysis of top-performing approaches on Duke, iLIDS, and Mars datasets. ``NL" represents a non-local block.}
    \scalebox{1.0}{
\begin{tabular}{l|cccccc}
\thickhline
\hline
\rowcolor[rgb]{0.92,0.92,0.92} {\cellcolor[rgb]{0.92,0.92,0.92}} & {\cellcolor[rgb]{0.92,0.92,0.92}} & \multicolumn{2}{c}{\textbf{MARS }} & \multicolumn{2}{c}{\textbf{DukeV }} & \textbf{iLIDS} 
\\
\rowcolor[rgb]{0.92,0.92,0.92} \multirow{-2}{*}{{\cellcolor[rgb]{0.92,0.92,0.92}}\textbf{Method }} & \multirow{-2}{*}{{\cellcolor[rgb]{0.92,0.92,0.92}}\textbf{Backbone }} & \textbf{mAP} & \textbf{R-1} & \textbf{mAP} & \textbf{R-1} & \textbf{R-1} \\
\hline
                                  
STAN\textsubscript{(CVPR'18)}     & Res-50     & 65.8 & 82.3 & $\times$    & $\times$     & 80.2  \\
Snippet\textsubscript{(CVPR'18)}  & Res-50     & 76.1 & 86.3 & $\times$    & $\times$     & 85.4  \\
STA\textsubscript{(AAAI'19)}      & Res-50     & 80.8 & 86.3 & 94.9 & 96.2  & $\times$     \\
ADFD\textsubscript{(CVPR'19)}     & Res-50     & 78.2 & 87.0 & $\times$    & $\times$     & 86.3  \\
VRSTC\textsubscript{(CVPR'19)}    & Res-50     & 82.3 & 88.5 & 93.5 & 95.0  & 83.4  \\
GLTR\textsubscript{(ICCV'19)}     & Res-50     & 78.5 & 87.0 & 93.7 & 96.3  & 86.0  \\
COSAM\textsubscript{(ICCV'19)}    & SERes-50   & 79.9 & 84.9 & 94.1 & 95.4  & 79.6  \\
%STT\textsubscript{(CVPR'18)}      & Res-50     & 86.3 & 88.7 & 97.4 & 97.6  & 78.0  \\
STE-NVAN\textsubscript{(BMVC'19)} & Res-50-NL  & 81.2 & 88.9 & 93.5 & 95.2  & $\times$     \\
MG-RAFA\textsubscript{(CVPR'20)}  & Res-50     & 85.9 & 88.8 & $\times$    & $\times$     & 88.6  \\
MGH\textsubscript{(CVPR'20)}      & Res-50-NL  & 85.8 & 90.0 & $\times$    & $\times$     & 85.6  \\
STGCN\textsubscript{(CVPR'20)}    & Res-50     & 83.7 & 90.0 & 95.7 & 97.3  & $\times$     \\
TCLNet\textsubscript{(ECCV'20)}   & Res-50-TCL & 85.1 & 89.8 & 96.2 & 96.9  & 86.6  \\
AP3D\textsubscript{(ECCV'20)}     & AP3D          & 85.1 & 90.1 & 95.6 & 96.3  & 86.7  \\
AFA\textsubscript{(ECCV'20)}      & Res-50     & 82.9 & 90.2 & 95.4 & 97.2  & 88.5  \\
HMN\textsubscript{(TCSVT'21)}     & Res-50     & 88.8 & 89 & 95.1 & 96.2  & $\times$  \\
SANet\textsubscript{(TCSVT'21)}     & Res-50     & 86.0 & 91.2 & 96.7 & 97.7  & $\times$  \\
DPRAM\textsubscript{(TIP'21)}     & Res-50     & 83.0 & 89.0 & 95.6 & 97.1  & $\times$  \\
PSTA\textsubscript{(ICCV'21)}     & Res-50     & 85.8 & 91.5 & 97.4 & 98.3  & 91.5  \\
STRF\textsubscript{(ICCV'21)}     & Res-50     & 86.1 & 90.3 & 96.4 & 97.4  & 89.3  \\
DenseIL\textsubscript{(ICCV'21)}  & Res-50     & 87.0 & 90.8 & 97.1 & 97.6 & 92.0  \\
STMN\textsubscript{(ICCV'21)}      & Res-50     & 83.7 & 89.9 & 94.6    & 96.7  & 80.6  \\
GRL\textsubscript{(ICCV'21)}      & Res-50     & 84.8 & 91.0 & $\times$    & $\times$     & 90.4  \\
TMT\textsubscript{(arXiv'21)}      & Res-50     & 85.8 & 91.2 & $\times$    & $\times$     & 91.3 \\ 

\thickhline
\end{tabular}
}
    \label{tab:Performance_Analysis}
    %\vspace{-10pt}
\end{table*}
\subsection{Triplet Loss}
\cite{chen2019spatial} design triplet loss to conserve ranking relationship among videos of pedestrian triplets. In triplet loss, the distance between feature pairs belonging to similar classes decreases, while the distance between feature pairs of different classes increases.
\begin{equation}
    \begin{aligned}
&L_{t r i}=\sum_{i, j, k \in \Omega}\left[d_{g}(i, j)-d_{g}(i, k)+m_{g}\right]_{+} \\
&+\sum_{i, j, k \in \Omega} \lambda\left[d_{l}(i, j)-d_{l}(i, k)+m_{l}\right]_{+}
\end{aligned}
\end{equation}
where $m_g$ and $m_l$ represent thresholds margin to restrict the distance gap between positive and negative samples and $[x]^+$ is the max function max$(0, x)$.
\subsection{Regressive Pairwise Loss (RPL)}
\cite{liu2018hierarchical} develop RPL to improve pairwise similarity by combining all positive sets in one single subspace. It helps with the soft margin between positive sets and is harder than the general triplet loss.
\begin{equation}
    \begin{gathered}
    L_{p}\left(x_{i}, x_{j}, y\right)=y \cdot \max \left\{d\left(x_{i}, x_{j}\right)-\log (\alpha), 0\right\} \\
    \quad+(1-y) \cdot \max \left\{\alpha-d\left(x_{i}, x_{j}\right), 0\right\}
    \end{gathered}
\end{equation}

where $y$ denotes label whether $x_i$ and $x_j$ are similar people. If a person is from the same identity it is represented as $1$ otherwise $0$. When y = 0, RPL pushes samples far away from
each other beyond margin $\alpha$. When y = 1, RPL pulls the samples together within distances no more than log($\alpha$).

%Please ignore it
\begin{comment}
\begin{table}[t]
    \centering
     \renewcommand\arraystretch{1.8}
     \setlength{\tabcolsep}{1.0mm}
    \caption{Specific loss functions for video re-ID.}
    \input{tables/loss_functions}
    \label{tab:Loss_Functions}
    \vspace{-5pt}
\end{table}
\end{comment}

\section{Datasets and Metrics}
We first describe the statistics of benchmark datasets that are frequently used for evaluating video re-ID methods. Secondly, we broadly review the performance of previous superior methods in chronological order. Lastly, we analyze results based on several major factors for video re-ID. 
%\vspace{-6pt}
\subsection{Training and Testing Datasets} Since video re-ID is a real-world problem and closer to a video surveillance scenario. During past years, various demanding datasets have been constructed for video re-ID: MARS \cite{zheng2016mars}, DukeMTMC-VID \cite{wu2018exploit} and iLIDS-VID\cite{wang2014person}, these three datasets are commonly used for training and evaluation, because of the large number of track-lets and pedestrian identities. 
%\vspace{-2pt}
\subsubsection{MARS} The dataset is constructed based on six synchronized CCTV cameras. It comprises $1,261$ pedestrians with different varieties of images (poor image quality, poses, colors, and illuminations) captured by two cameras. It is extremely difficult to match pedestrian images because it contains $3,248$ distractors to make the dataset more real-world.
%\vspace{-2pt}
\subsubsection{DukeMTMC-VID} It is a subgroup of the DukeMTMC dataset which purely consists of 8 cameras with high-resolution images. It is one of the large-scale datasets where pedestrian images are cropped using manual hand-drawn bounding boxes. Overall, it comprises $702$ identities, $16,522$ training images $17,661$ gallery images, and $2,228$ probe images.
%\vspace{-2pt}
\subsubsection{iLIDS-VID} It is one of the challenging datasets which contains $300$ pedestrians captured by two CCTV cameras in public. Due to the public images, it contains lighting, viewpoint changes, different similarities, background clutter, and occlusions. It consists of a $600$ sequence of images of $300$ diverse individual images. Each sequence of the pedestrian images has a range length of $23$ to $192$ and the number of frames is $73$.

\subsection{Evaluation Protocol}

The are two standard evaluation protocols for evaluating video re-ID methods which are mAP and CMC. CMC is the probability of top top-K correct matches in a retrieval list. 
% In most of the cases, the IoU value with GTs is greater than or equal to 0.5. 
Another evaluation metric is mAP, which measures the average retrieval accuracy with multiple GT.

\section{Analysis and Future Direction}
We broadly review the top-performing methods from video re-ID perspectives. We mostly focus on the work published in $2018$ till now. Specifically, we include STAN \cite{li2018diversity}, Snippet \cite{chen2018video}, STA \cite{fu2019sta}, ADFD \cite{zhao2019attribute}, VRSTC \cite{hou2019vrstc}, GLTR \cite{li2019global}, COSAM \cite{subramaniam2019co}, STE-NVAN \cite{liu2019spatially}, MG-RAFA \cite{zhang2020multi}, MGH \cite{yan2020learning}, STGCN \cite{yang2020spatial}, TCLNet \cite{hou2020temporal}, AP3D \cite{gu2020appearance}, AFA \cite{chen2020temporal}, PSTA\cite{wang2021pyramid} DenseIL \cite{he2021dense}, STMN \cite{eom2021video}, STRF \cite{aich2021spatio}, SANet \cite{bai2021sanet}, DPRAM \cite{yang2021two}, HMN \cite{wang2021robust}, GRL \cite{liu2021watching}, and TMT \cite{liu2021videoTTTrigeminal}. We summarize the video re-ID results on three widely used benchmark datasets. Table. \ref{tab:Performance_Analysis} highlights the backbone, mAP and R-1 results, and methods.

Firstly, with the recent development of self-attention-based methods, several video re-ID methods have obtained higher mAP and top-1 accuracy (\cite{liu2021videoTTTrigeminal} 91.2\%) on the widely used MARS dataset. Especially, DenseIL \cite{he2021dense} achieves the highest mAP of 87.0\% but rank-1 accuracy is 90.8\% which is slightly lower than TMT\cite{liu2021videoTTTrigeminal} on MARS dataset. The advantage of the DenseIL \cite{he2021dense} method is to simultaneously use CNN and attention-based architecture to efficiently encode spatial information into discriminative features. Those methods focus on long-range relationships and specific part-level information on an input signal. Various popular methods separately learn weights and spatial-temporal features \cite{hou2019vrstc,hou2020temporal}. Another observation in \cite{zhang2021spatiotemporal} illustrates that capturing and aggregating pedestrian cues is spatial-temporal while ignoring discrepancies including background areas, viewpoint, and occlusions. However, in a real-world scenario, the visual data contains a lot of diverse modalities such as recording information, camera ID, etc. Most studies focus on visual similarity by matching probe images into gallery images. Thus, it neglects textual information which is not a good idea. Proposing a new method that extracts visual-textual information at the same time would be helpful in a real-world environment and it will also help to provide more accurate results.
\par
Secondly, annotating new datasets with accurate labels on different CCTV cameras is an expensive and laborious task. In most cases, annotated data are wrong-labeled due to various factors such as person visibility, background clutter, and noise issues in images. Several researchers focus on unsupervised methods \cite{ye2018robust,ye2019dynamic} and active learning approaches \cite{wang2018deep} to alleviate the annotation problem. Still, the accuracy of unsupervised video re-ID methods degrades significantly compare to supervised video re-ID methods. In the future, introducing a unique video re-ID method that facilitates clustering and label assignment will be considered to improve existing unsupervised methods. Further, designing a specific data augmentation policy in a re-ID search space can easily increase the overall performance for all re-ID methods.
\par
Finally, the accuracy on three challenging datasets reaches a difficult state, where the performance gap is less than 1\% accuracy like PSTA \cite{wang2021pyramid} and DenseIL \cite{he2021dense} on the DukeVID dataset. As a result, it is still difficult to select the best superior method. On iLIDS, the rank-1 performance of PSTA \cite{wang2021pyramid} is 91.5\% and TMT \cite{liu2021videoTTTrigeminal} is 91.3\%. However, most video re-ID architectures are complex in terms of the number of parameters for learning invariant feature representations on combined datasets. Meanwhile, re-ID methods use metric learning techniques like euclidean distance to calculate feature similarity which is time-consuming and slow retrieval and not applicable in real-world applications. How to design a new strategy to replace metric learning strategies still needs more research. Thus, further exploration of video re-ID approaches remains an interesting area for future research.
\section{Conclusion}
This paper presents a comprehensive review of global appearance, local part alignment methods, graph learning, attention, and transformer model in video re-ID. We provide specific loss functions with mathematical representation to help new researchers to use them instead of using straightforward common loss functions for video re-ID. Finally, we highlight widely and frequently used datasets for evaluating video re-ID techniques and analyze the performance of different methods and provide future research direction.

\bibliography{ijcai22.bib}
\bibliographystyle{plain}

\end{document}